\documentclass[letterpaper, 10 pt, conference]{ieeeconf}

\IEEEoverridecommandlockouts                              % This command is only needed if 
\usepackage{amsmath} % assumes amsmath package installed
\usepackage{amssymb}  % assumes amsmath package installed
\usepackage{siunitx} % for unit http://texdoc.net/texmf-dist/doc/latex/siunitx/siunitx.pdf
\usepackage{graphicx}
\usepackage{url}
\usepackage[ruled]{algorithm2e}% http://ctan.org/pkg/algorithm2e
\graphicspath{{Figures/}}

\usepackage[dvipsnames]{xcolor}        % use colored text
\title{\LARGE \bf
A Tunably Compliant Origami Mechanism for Dynamically Dexterous Robots
}

\author{Wei-Hsi Chen$^{1}$, Shivangi Misra$^{4}$, Yuchong Gao$^{2}$, Young-Joo Lee$^{1,2}$,\\ Daniel E. Koditschek$^{1}$, Shu Yang$^{2}$, Cynthia R. Sung$^{3}$% <-this % stops a space
\thanks{$^{1}$Department of Electrical and Systems Engineering,
$^{2}$Department of Materials Science and Engineering,
$^{3}$Department of Mechanical Engineering and Applied Mechanics,
$^{1234}$University of Pennsylvania, General Robotics, Autonomous, Sensing and Perception (GRASP) Lab}
}

\begin{document}

\maketitle
\thispagestyle{empty}
\pagestyle{empty}

%%%%%%%%%%%%%%%%%%%%%%%%%%%%%%%%%%%%%%%%%%%%%%%%%%%%%%%%%%%%%%%%%%%%%%%%%%%%%%%%
\begin{abstract}
We present an approach to overcoming challenges in dynamical dexterity for robots through tunable origami structures. Our work leverages a one-parameter family of flat sheet crease patterns that folds into origami bellows, whose axial compliance can be tuned to select desired stiffness.  Concentrically arranged cylinder pairs reliably manifest additive stiffness, extending the tunable range by nearly an order of magnitude and achieving bulk axial stiffness spanning 200--1500 N m$^{-1}$ using 8 mil thick polyester-coated paper. Accordingly, we design origami energy-storing springs with a stiffness of {1035~N m$^{-1}$} each and incorporate them into a three degree-of-freedom (DOF) tendon-driven spatial pointing mechanism %whose conventionally actuated arrangement of three parallel tendon-driven concentric origami cylinders 
that exhibits trajectory tracking accuracy less than 15\% rms error within a  ($\sim$2 cm)$^3$ volume. The origami springs can sustain high power throughput, enabling the robot to achieve asymptotically stable juggling for both highly elastic (1~kg resilient shotput ball) and highly damped (``medicine ball'') collisions in the vertical direction with apex heights approaching 10 cm. % and a highly damped ``medicine ball'' of half that  mass to  half that height.
%To the best of our knowledge this is the first 
The results demonstrate that ``soft'' robotic mechanisms %in the reported literature to 
are able to perform a controlled,  dynamically actuated task.  
\end{abstract}

%%%%%%%%%%%%%%%%%%%%%%%%%%%%%%%%%%%%%%%%%%%%%%%%%%%%%%%%%%%%%%%%%%%%%%%%%%%%%%%%
\section{Introduction}

Over decades of  robot manipulation  \cite{Buehler_Koditschek_1987} and locomotion~\cite{Buehler_Koditschek_1988} research, the term {\em dynamical dexterity}   has come to mean the programmed   \cite{Burridge_Rizzi_Koditschek_1996}  exchange  of  work and information  at high temporal rates~\cite{Rizzi_Koditschek_1996}. Indeed,  sensorimotor dexterity \cite{Venkadesan_Guckenheimer_Valero-Cuevas_2007} is essential to the  quality of our daily life \cite{Lawrence_Fassola_Werner_Leclercq_Valero-Cuevas_2014}, specifically in the high-strength regime \cite{Valero-Cuevas_Smaby_Venkadesan_Peterson_Wright_2003}. As robots begin to enter the unstructured workplace, their users' expectation of companionable dexterity will continue to sharpen the intrinsic conflict between the need for more actuated degrees of freedom and the requirement of high power density  \cite{revzen2017we}, whose limits in the relevant highly energetic and high strength regime  have long manifested as the first scarce resource in conventional robot actuation technologies \cite{Hunter_Hollerbach_Ballantyne_1992}.

%\subsection{Related Literature}

The use of soft materials of varied shape and tunable compliance enjoys an active literature in contemporary robotics~\cite{laschi2014soft} and beyond~\cite{howell2001compliant} as a method for introducing both high maneuverability and resilience directly into the body of a robot.
However, while compliant elastomeric robots have have occasionally been demonstrated to produce  fast, dynamic~\cite{marchese2014autonomous}, and even explosive~\cite{tolley2014untethered} maneuvers, the high damping and high fatigue properties in these elastomers often limit these maneuvers to a single use. Meanwhile, sustained dynamic motions needed for tasks such as juggling, hopping, and trotting, remain out of reach for most soft robots.

Origami-inspired approaches to replacing~\cite{6266749,jones2006,mcmahan2005} or enhancing~\cite{8794068} soft-bodied machines promise to address these challenges in achieving repeated, dynamic movement.
%To this end, we fabricate
Past research in this field has demonstrated durable actuators from origami cylinders, yielding  lightweight structures~\cite{hoover2008roach,whitney2011} patterned by high compliance folds.
%Origami robots have recently demonstrated comparable high degree of freedom behavior~\cite{jeong2018design}, with the added advantages of low weight, low fatigue, and simplified fabrication and assembly on the order of hours~\cite{hoover2008roach}.
 The resulting actuators assert high specific force~\cite{martinez2012elastomeric} over a large volume-to-mass  workspace \cite{zhang2016extensible}, and bear substantial loads \cite{8794068} while resisting unwanted (e.g.,  torsional)  disturbances \cite{8206027}.
%A  general  trend  among  these  demonstrations  is to  use  the  origami  pattern  to  induce  kinematic  constraints (i.e.,  directionality  of  motion)  and  rely  on  the  folds  or additional  attachments  to  generate  the  desired  stiffnesses.
However, to date, origami robots have been designed as though with  rigid linkages joined through rotational folds, without taking into consideration of the additional compliance and resiliency provided by the sheet material itself.
As a result, they have been unable to match the \textit{power} densities of the rigid-body counterparts.

In this paper, we explore the prospects for integrating tunable compliance and highly energetic anisotropic %(i.e. simultaneously directionally ``rigid'' and ``soft'') 
designs in the drive  train of a three DOF robotic limb through the lens of the vertical one-juggle \cite{Buehler_Koditschek_1987}, a well established route toward dynamically dexterous manipulation and locomotion \cite{Johnson_Koditschek_2013}.
Through geometric designs of an origami bellow pattern~\cite{REBO2018}, we aim to achieve  elastic axial compliance 
with reduced material weight and mitigation of energetic loss,
thus producing a ``soft" spring robotic juggler capable of  high-power operation.
%It takes explicit advantages of the soft-rigid hybrid nature in an origami design to achieve high-force, high-deformation motion while minimizing fatigue and energy loss. 
It has been suggested in origami mechanics~\cite{reid2017geometry,filipov2015origami} that the resistance of an origami design to static loading conditions can be tuned through appropriate choices of its geometric parameters.
Here, we demonstrate that the \textit{dynamic} response of an origami pattern can also be tuned, and that the resulting structure is in fact capable of transducing the high power densities required for dynamical dexterity.

Specifically, we leverage the Reconfigurable Expanding Bistable Origami (REBO) pattern~\cite{REBO2018}, which was originally designed for geometric reconfiguration. 
Interestingly, we find that small changes in the fold pattern alter not only the geometry of the structure, but also its rigidity. 
Thus, we  manipulate the REBO design parameters for a dynamic juggling task and
%We exercise a geometric fold parameter of the REBO architecture to achieve systematically tunable axial compliance while
introduce a concentric pairing of the REBO cylinders to enhance stiffness.  We drive three such concentrically paired cylinders to achieve \SI{3e3}{\newton\per\metre} stiffness and compress each via a conventionally actuated tendon. The resulting three-DOF ``limb" achieves reasonably good trajectory tracking (less than 15\% rms error) within a workspace whose actuated volume is limited to a small fraction of its kinematically achievable span by the torque output of the brushless DC motors. Nevertheless, this volume affords adequate travel and the paired REBO cylinders transduce sufficiently high power to achieve asymptotically stable vertical juggling of balls of varied mass and resilience.      

%%%%%%%%%%% Hopefully this has already been covered in the preamble %%%%%%%%%%%
%To demonstrate that REBO-mediated actuation is able to perform dynamical tasks, a juggling task is assigned to the REBO Juggler, where it is required to repeatedly hit a weighted ball into the air with its top plate as the paddle. Stable juggling is a quintessentially dynamical task in the sense of requiring carefully controlled exchanges of potential and kinetic energy between a robot and its environment \cite{buhler1990family}. The task is an effectively simple but appropriately challenging precursor to more general dynamically dexterous capabilities because the duration of travel during paddle-ball contact is small while the force required to change the direction of the motion of the ball is substantial.

%%%%%%%%%%%%%%% Add this to the conclusion if space permits %%%%%%%%%%%%%%%%%%%%
%% WC: I'll move this back to section V
%Reflecting the past traditions of the field of dynamical locomotion, the core problem of stable running \cite{raibert1986legged} can reduced to the problem of stable vertical hopping \cite{raibert1983dynamic}, for which a purely vertical juggle is representative surrogate \cite{buehler1990simple,koditschek1991analysis}. Thus, as a proof of concept task, the REBO juggler’s ball, its dynamical ``environment", is restricted to purely vertical motion by having the ball bounce inside a tube mounted on the paddle, while the three parallel REBO can be viewed as one virtual actively loaded vertical spring. 
%%%%%%%%%%%%%%%%%%%%%%%%%%%%%%%%%%%%%%%%%%%%%%%%%%%%%%%%%%%%%%%%

In summary, the contribution of this work is the development, analysis, and application of a new approach for dynamically dexterous  manipulation; it substitutes an origami structure for a conventional spring, storing sufficient energy and transducing it with sufficient power and force to juggle stably  a 1 kg mass to a height selectable over a range of nearly 10 cm, from initial conditions within a simlarly large basin of attraction.    
%The paper is organized as follows. Section II describes the origami pattern design and its material properties. Section III introduces the juggler robot composed of REBO and a kinematic model description of it. Section IV shows the result of the robot as a pointing device. Section V discusses the experimental results performing the juggle task. Section VI concludes with directions for future work.

\section{Origami Module Design}
\subsection{Parameterized Programmable Crease Pattern}
    \begin{figure}[t]
      \centering
      \includegraphics[scale=0.58]{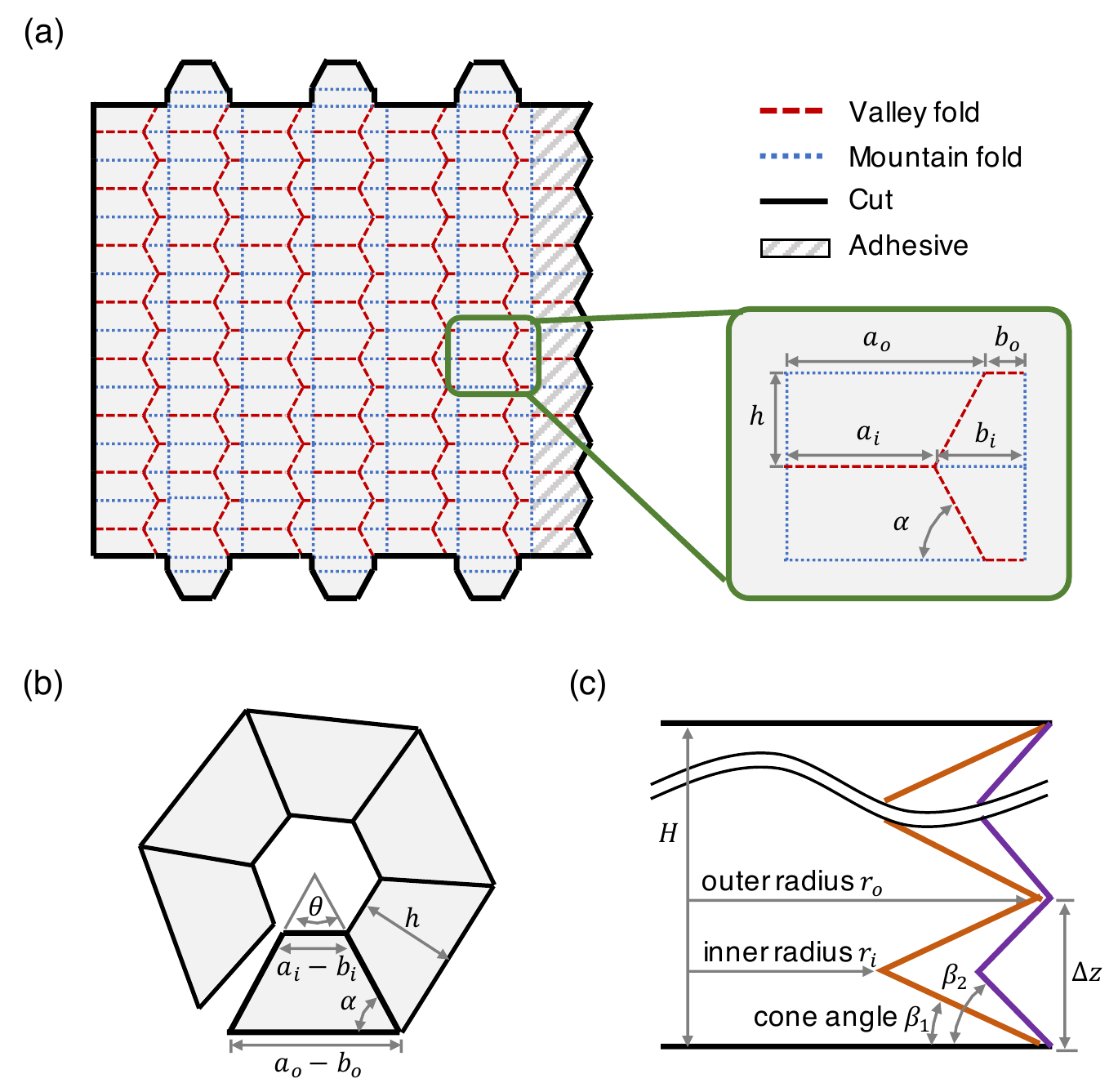}
      \caption{Origami REBO design (a) the crease pattern of REBO (b) schematic diagram of one layer of REBO when folded flat (c) schematic diagram of the cross section of the right half of the the double layer REBO structure, where the purple and orange line indicates different layers. $\beta$ denotes the cone angle, where the index indicates the different layer}
      \label{fig1}
    \end{figure}
    
%We are interested in understanding the role of soft structures with programmable compliance in robot dexterity. 

The REBO design (Fig.~\ref{fig1}(a)) is an origami bellows.
The fold pattern is a tessellation of rectangular units arranged into $n_r$ columns and $n_l$ rows, with 
the left and right columns glued together to form a tube.
Each unit contains a middle crease at an angle  $\alpha$ from horizontal, as shown in the grey box in Fig. \ref{fig1}(a). 
When folded, these creases cause each row of the structure to collapse into a $n_r$-sided right frustum with height 
%When folded, the REBO is a piecewise-linear bellows structure, where the conical surface with height
$\frac{1}{2}\Delta z$ and side lengths $(a_o-b_o)$ and $(a_i-b_i)$ on the larger and smaller bases (Fig.~\ref{fig1}(b)). %The side length  of the regular $n_r$-sided polygon at the base is $(a_o-b_o)$, and 
We define the angle between the base and  side of the layer as the cone angle $\beta$.

There is a direct relationship between the geometric parameters of the fold pattern and those of the 3-D folded state.
In particular, the rotation angle $\theta$ of each trapezoid shown in Fig.~ \ref{fig1}(b) is $\theta=(2 \pi \cos {\beta})/n_ {r}$. The values of $\alpha$ and $h$ can then be calculated as
\begin{align}
    \alpha &= (\pi -\theta)/2\\
    h &= \Delta z \csc{\beta}.
\end{align}
%and \WC{!!would not or would slightly} change the mechanical property

The design has the ability to store potential energy in the bending of the folds and the stretching of the faces, similarly to the multistable ``bendy straw'' design~\cite{bende2018overcurvature}.
Interestingly, by changing the size of the design and geometric parameters such as $\beta$ and $\Delta z$, the amount of structural deformation required for the design to bend and compress can be manipulated, thus allowing us to control the stiffness of the design purely through its geometry.
%``Bendy straw", or a cylindrical bellow structure with alternating conical surface, has the ability to store potential energy in the folds as well as the surface region when compressed. While \cite{bende2018overcurvature} focuses on the multistability of the structure, this paper tries to explore the elastic deformation of the structure and consider it a spring by presenting an origami structure based on this bellow structure. 
%Once a desired cone angle $\beta$ is found with respect to the stiffness, as further discussed in the next section, the crease pattern could be generated and fabricated. The crease pattern is composed of $n_r$ units in each row, with $n_l$ layers. 
When $\beta$ is $0$, the folded state 
is a flat polygon with little resistance to axial forces. This is because the flat folded configuration  relies on torsional stiffness in the folds, which is typically small. As $\beta$ increases, the slope of each layer increases, and REBO cannot be folded flat without deformation of the fold surfaces. In other words, the potential energy of compressing the structure is now stored not only in the folds, but also in the sheet surfaces, making the structure stiffer. Thus by designing $\beta$, one can generate spring-like structures with variable degree of stiffness.

\subsection{Effect of Cone Angle on Stiffness}
    \begin{figure}[t]
      \centering
      \includegraphics[scale=0.58]{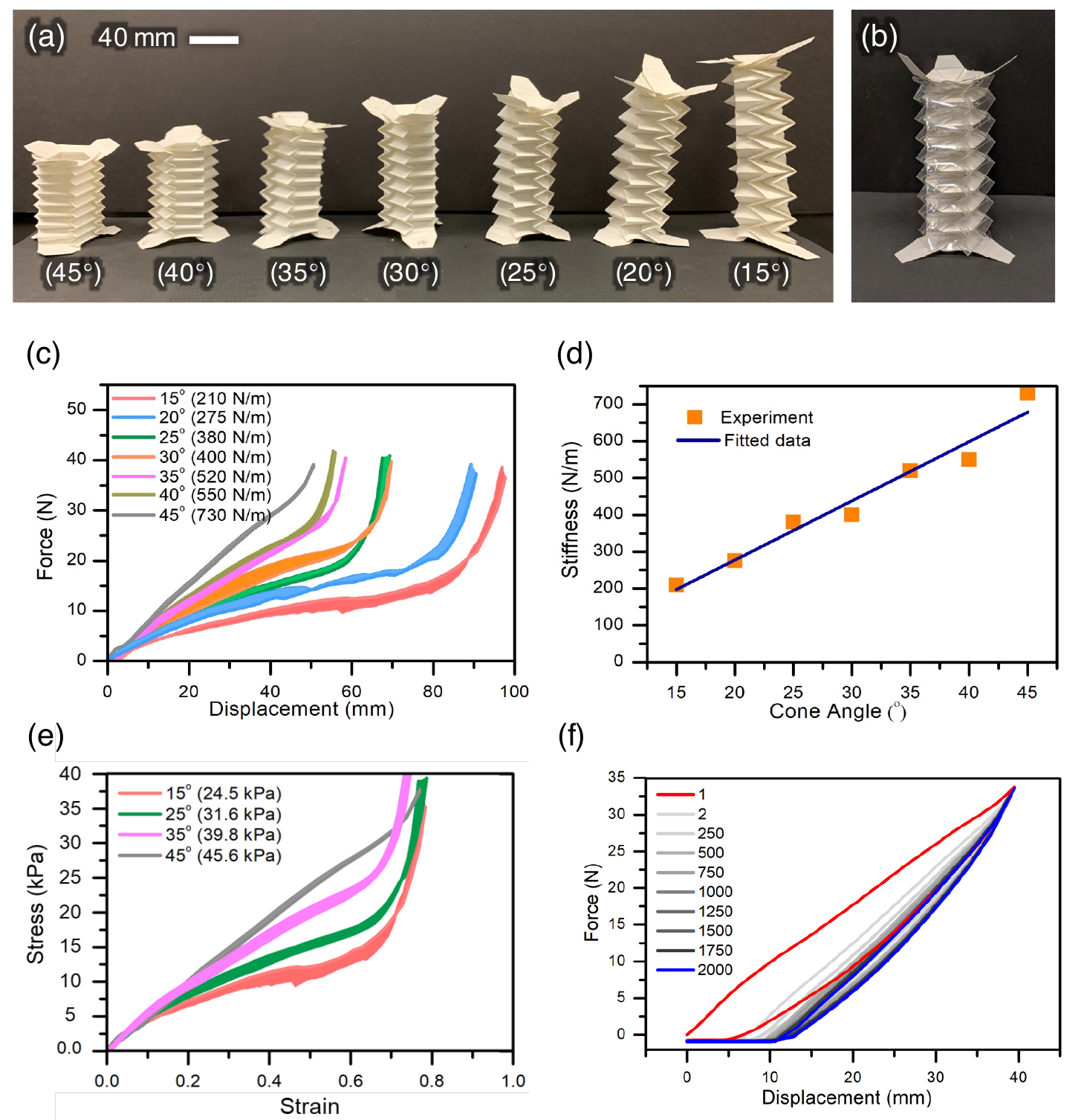}
      \caption{Material performance, stiffness vs. cone angle: (a) specimens of different cone angles  (b) the double layer REBO structure with a transparent outer layer and a white inner layer. (c) compression force vs. displacement (d) linear fit of the stiffness experiments (e) stress vs. strain (f) cyclic test}
      \label{fig2}
    \end{figure}

To understand the  relation between the cone angle and the stiffness of the REBO, %can be found through a mechanical model based on experiments. 
we folded multiple versions of the pattern with variable geometric parameters and conducted compression tests.
All samples were folded out of 8 mil thick Durilla synthetics paper with polyester finish (CTI Paper, USA) and used  3M 467MP adhesive transfer tape to glue the left and right sides together as a closed cylinder. The parameters chosen for this study were as follows: ${a_o=\SI{20}{\milli\metre}}$, $b_o =\SI{6}{\milli\metre}$, $\Delta z=\SI{10}{\milli\metre}$, $n_r = 6$, and a total of $n_l=8$ layers. We tested cone angles $\beta$ between \ang{15} and \ang{45}, with increments of \ang{5} in between. The crease patterns are then generated from a MATLAB script. Fig.~\ref{fig2}(a) shows the results of the fabrication. The theoretical rest length of all of these REBO structures should be $l_{t}=n_l \Delta z = \SI{80}{\milli\metre}$. However, due to the imperfection of manual folding, the final rest lengths are not equal, and in fact decrease as $\beta$ increase. The true rest lengths $l_{real}$ were measured and used for the following experiments.

%A compression test is performed on all these specimen to establish a physical model of the relation between the cone angle and the stiffness.
We used an Instron Model 5564 with 100N compression load cell to measure the force required to compress each specimen from its natural length $l_{real}$ until all the layers were stacked flat and the force exceeded 40N. %with a force sensor mounted on it. The displacement after the load cell starts to turn nonzero is measured along with the force data. 
Each specimen was measured 5 times.

The result of the experiment is shown in Fig.~\ref{fig2}(c). 
The shaded regions show the minimum and maximum  force values corresponding to each displacement for the sample.
The results show that REBO structure exhibits a Hookean force-displacement curve for well over 2/3 of its total travel. We computed the effective stiffness (elastic constant) as the slope of the linear fit in this region. 
The sharp increase in stiffness at the end of the curve corresponds to all of the layers coming into contact with each other so that the specimen can be viewed as a solid cylinder. %, and the slope of this region is the Young's Modulus of the paper itself
% \AK{I'm not quite sure if the latter half of the sentence is true, It's certainly not Young's modulus since we are doing force-displacement, and I don't think it's measuring the paper's property, it's basically the compression jigs pushing against each other, can we delete this or change it to ' all of the layers coming into contact with each other and can be view as an solid paper cylinder'?}. \CS{good?} \WC{looks good to me}
This region should be avoided during application.
Fig.~\ref{fig2}(d) summarizes the mean effective stiffness for each of the measured samples.
A linear fit indicates that the stiffness $K_s$ increases as a roughly affine function of the cone angle $\beta$, $K_s= 16.0571\beta - 43.7143$, with $R^2= 0.9548$.
%\CS{we would expect it to vary with sin(beta). sin is approximately linear in the domain we're considering. Not sure if it's worth mentioning here - you would need to explain.} \WC{ I think we can get away with just a linear fit, explaining sin would require the model itself, which is not presented in this paper.}
The REBO design was able to achieve a broad range of stiffnesses from \SIrange{210}{730}{\newton\per\metre}. 

We computed stress-strain curves for each of the specimens (Fig.~\ref{fig2}(e)).
%Fig. \ref{fig2}(e) shows the stress-strain curve of the specimens, where the 
Strain was calculated using the real rest length $l_{real}$ and the stress was the compression force acting on the effective hexagonal area $A=3\sqrt{3}(a_o-b_o)^2$,
with the Young's modulus as the slope of the resulting curve. %was computed as the slope of the linear region. %can be found from the figure. 
The results show %\AK{Young's modulus is not really dimensionless.}also shows
that the cone angle does indeed have a significant effect beyond simply changing the length of the REBO. % on the stiffness. 

During experiments, we found that the dimensions of the REBO affect the stress-strain profile. 
For REBOs with the same cone angle $\beta$, a larger side length and height reduces the Young's modulus. As a result, a higher $\beta$ is required to achieve a similar profile for a larger-scale model. %Theoretically, this work can be generalized to any material, and the stress-strain profile of REBO can be predicted for given size, thickness, and Young's modulus of the material. 
The scaling study of this REBO structure will be the focus of future work.
% The dimensionless result also shows that the cone angle does indeed have a significant effect to the change of stiffness. Thus by changing the geometry pattern of the structure, the same material can result in very different property (Young's Modulus).
% On the other hand, thickness of the paper also effects the stiffness in a cubical scaling term, as predicted in \cite{sung2015foldable}. 

\subsection{Double-Layered Design}

% \CS{I suggest deleting this paragraph. The measured curves in this paper don't really show any evidence of bistability, likely because of the material used. B=35 and B=45, which were used for the outer layers, look very linear, actually.}\WC{I'm fine with cutting it! the following paragraph has pretty detailed discription already} Due to the fact that $\beta\neq0$, REBO would collapse into a stable state when it is compressed till a point one of the conical surface of a single layer buckles too much, which is referred to the bistability of the structure. The bistability of REBO is useful in some quasistatic representation \cite{REBO2018}, but is not suitable for dynamical task. This research introduces a double layer REBO structure to get rid of the bistability, as shown in Fig. \ref{fig1}(c) and \ref{fig2}(b), where a REBO structure with a lower cone angle is enclosed in another structure with a higher one. The collapse of the outer structure can be prevented due to physical constraint of the inner paper. The double layer structure also provides higher stiffness since it could be viewed as two springs in parallel, and its stiffness is the sum of both springs'.

    \begin{table}[t]
        % \scriptsize
        \caption{Stiffness of Double-Layered REBO}
        \label{tab1}       % Give a unique label
        % Follow this input for your own table layout
        \begin{center}
        \begin{tabular}{c c | c}
         Inner layer ($\beta_{il}$)  & Outer layer ($\beta_{ol}$) & Double layer\\
        \noalign{\smallskip}\hline\noalign{\smallskip}
        \SI{320}{\newton\per\metre} (\ang{25}) & \SI{540}{\newton\per\metre} (\ang{35}) &  \SI{888}{\newton\per\metre} \\
        \SI{320}{\newton\per\metre} (\ang{25}) &  \SI{725}{\newton\per\metre} (\ang{45}) &  \SI{1035}{\newton\per\metre}\\
         \SI{700}{\newton\per\metre} (\ang{35}) &  \SI{725}{\newton\per\metre} (\ang{45}) &  \SI{1490}{\newton\per\metre}\\
        \end{tabular}
        \end{center}
    \end{table}

The results show that a maximum stiffness of \SI{730}{\newton\per\metre} for the REBO design is achieved at $\beta=\ang{45}$.
Above this $\beta$ value, the structure is at risk of buckling irreversibly upon compression. 
However, higher stiffness can be achieved by arranging multiple REBO structures concentrically, as shown in Fig.~\ref{fig1}(c) and~\ref{fig2}(b).
This parallel spring structure demonstrates additive stiffness and protects against snap-through buckling to the bistable inverted configurations, which were previously demonstrated in other applications \cite{REBO2018}. Here, three sets of double layer structure have been fabricated, with the cone angle of the inner and the outer structure to be $(\ang{25},\ang{35})$, $(\ang{25},\ang{45})$, and $(\ang{35},\ang{45})$, respectively. To facilitate the fabrication of the double layered REBO structure, we  increased $b_o$ to make room for the inner structure to slide through, then refolded it to enclose it. A compression test was performed before and after the combination and the stiffness of each specimen was measured (see summary of results in Table \ref{tab1}). The stiffness of the double layer structure is indeed the sum of the stiffness of the two individual layers with a maximum error of only $3.2\%$. It is to our observation that by increasing $b_o$, the stiffness decreases a little due to the fact that there is more space for the paper to deform.
    
    % \begin{figure}[t]
    %   \centering
    %   \framebox{\parbox{3in}{We suggest that you use a text box to insert a graphic (which is ideally a 300 dpi TIFF or EPS file, with all fonts embedded) because, in an document, this method is somewhat more stable than directly inserting a picture.}}
    %   %\includegraphics[scale=1.0]{figurefile}
    %   \caption{Maybe Finite Element modeling (\textcolor{purple}{YoungJoo eta 7/8}), or maybe one angle diff damping constant wrt 4 different cone angle? and double layer(\textcolor{purple}{Shivangi eta 7/15})}
    %   \label{fig2c}
    % \end{figure}

\subsection{Repeatability and Energy Loss}
Finally, for dynamic robot applications, it is important to understand the energy dissipation and resilience of the REBO design.
We therefore experimentally measured the response of a $\beta=\ang{45}$ REBO under  cyclic loading.
%A durability test is performed on a specimen with $\beta=\ang{45}$ for 2000 cycles of loading and unloading 
The specimen was alternately compressed and released between its original rest length and \SI{40}{\milli\metre} displacement for 2000 cycles. Each cycle took 12 seconds.
The results are shown in Fig. \ref{fig2}(f).

While the first cycle (red) required much higher forces than the rest of the trials, the response of the REBO quickly converged to the blue curve after the second trial, and remained consistent for the rest of the trials. This behavior is consistent with literature in origami mechanics, where the first folding is often an outlier~\cite{qiu2013kinematic} since it plastically deforms the material and changes the structure's equilibrium state.

For repeated dynamic tasks, we are primarily concerned with the steady-state behaviors, i.e., the blue curve.
We observed  elastic hysteresis between the tension and compression portions of the tests, suggesting that more energy was required for loading comparing to unloading, and thermal energy was dissipated during this process. 
%We also observe a large change in the curve between the first (red) and second tests, and a gradual shift in response until it converges to the blue curve.
%As the durability test progress, the  curve shifts slowly; when the same force is applied in both the loading and unloading phase, the displacement gets larger with cycles. 
The displacement offset after 2000 runs is small compare to the original length, with a maximum of \SI{3}{\milli\metre}, $5\%$ of the rest length. After 2000 cycles, no physical damage was observed on the specimen and no failure was found on the force-displacement plot.

\section{Juggling Robot Design}

Our characterization shows that stiffness on the order of \SI{e3}{\newton\per\metre}, the range where energy exchange with \SI{e0}{\kilogram} loads has been shown to achieve useful aerial-phase compliant-legged running gaits \cite{galloway2007design}, can be easily accessible, and importantly that this performance does not degrade over repeated uses. Armed with the understanding, we integrate the programmable compliance design to legged robots in the form of the ``REBO Juggler." Juggling a weighted ball continuously at a certain height requires a periodic motion and great power, similar to many dynamical locomotion tasks.
%\CS{need to motivate your design choices here, particularly in relation to the characterization in the prev 3 pages - how is this?}\WC{actually, yeah, this is the right reason}

\subsection{Robot Platform}
    \begin{figure}[t]
      \centering
      \includegraphics[scale=0.58]{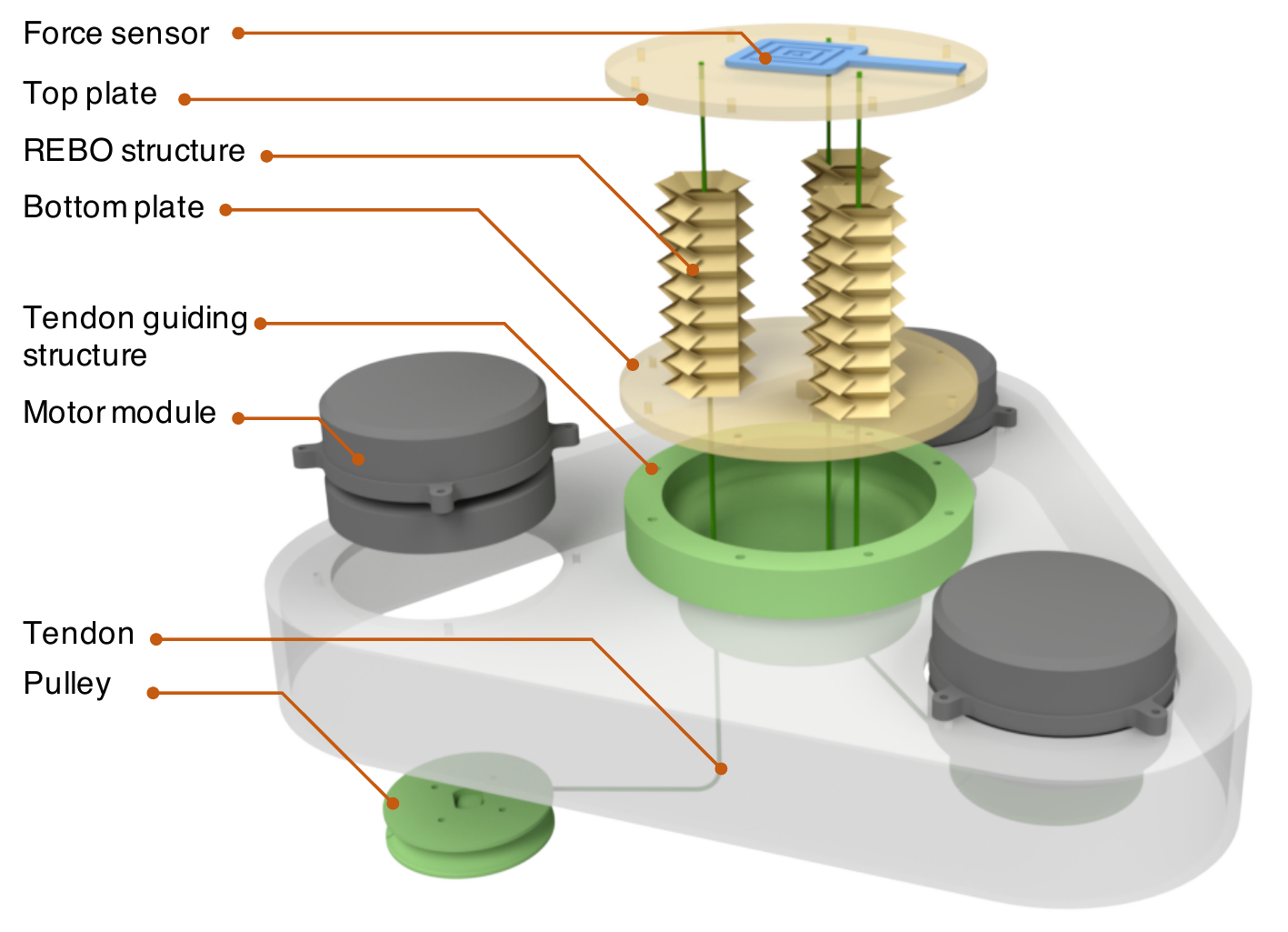}
      \caption{CAD rendering of the REBO Juggler}
      \label{fig3}
    \end{figure}

%When designing legged robot for dynamical locomotion, force is not the only character to be considered when choosing an actuator, but also power. The duration where the work is done by the leg to the ground is often very small, thus large power is required. This is the main obstacle for utilizing soft robotics mechanism to legged robots. Here this work purpose a robot platform, 
 %, an example of a task that requires great power, using the purposed REBO structure as a part of the actuator.
The robot (Fig.~\ref{fig3}) consists of four main parts: (a)~the compliant REBO body, (b)~force transmitting system composed of brushless DC electrical motor modules (Ghost Robotics MNSB01 Sub-Minitaur U8 Motor Module~\cite{ghostrobotics}) with tendon (Sufix 832 Advanced Superline Braid) and 3D-printed pulley system, (c)~contact detection using force sensor (Ohmite FSR01CE), and (d)~a microprocessor (Ghost Robotics MNS043 mainboard \cite{ghostrobotics}) for integrating sensing and control.
The compliant body, shown in Fig. \ref{fig4}(a), is composed by three double-layered REBO structures with a stiffness of \SI{1035}{\newton\per\metre} each. The parameters of the outer layer are $\beta = \ang{45}$, $a_o=\SI{20}{\milli\metre}$, $b_o=\SI{5}{\milli\metre}$, $\Delta z = \SI{10}{\milli\metre}$, $n_r = 6$, $n_l = 8$, and the ones for the inner layer are $\beta = \ang{25}$, $a_o=\SI{19}{\milli\metre}$, $b_o=\SI{0}{\milli\metre}$, $\Delta z = \SI{10}{\milli\metre}$, $n_r = 6 $, $n_l = 8$. Each REBO weighs about \SI{16}{\gram}. Three REBOs were then mounted between the top and bottom acrylic plates and secured using tabs, forming the compliant body that weighs about \SI{220}{\gram}. The tendon was laced through the structural through-holes of REBO, with one end fixed on the top plate and the other on the pulley mounted on the motor. Rotating the motor compressed or released the REBO, and the speed limitation of the linear motion on REBO was determined by the motor. A force sensor was placed on top of the top plate to detect when an object was in contact with the top plate.

\subsection{Kinematic Model}
    \begin{figure}[t]
      \centering
      \includegraphics[scale=0.58]{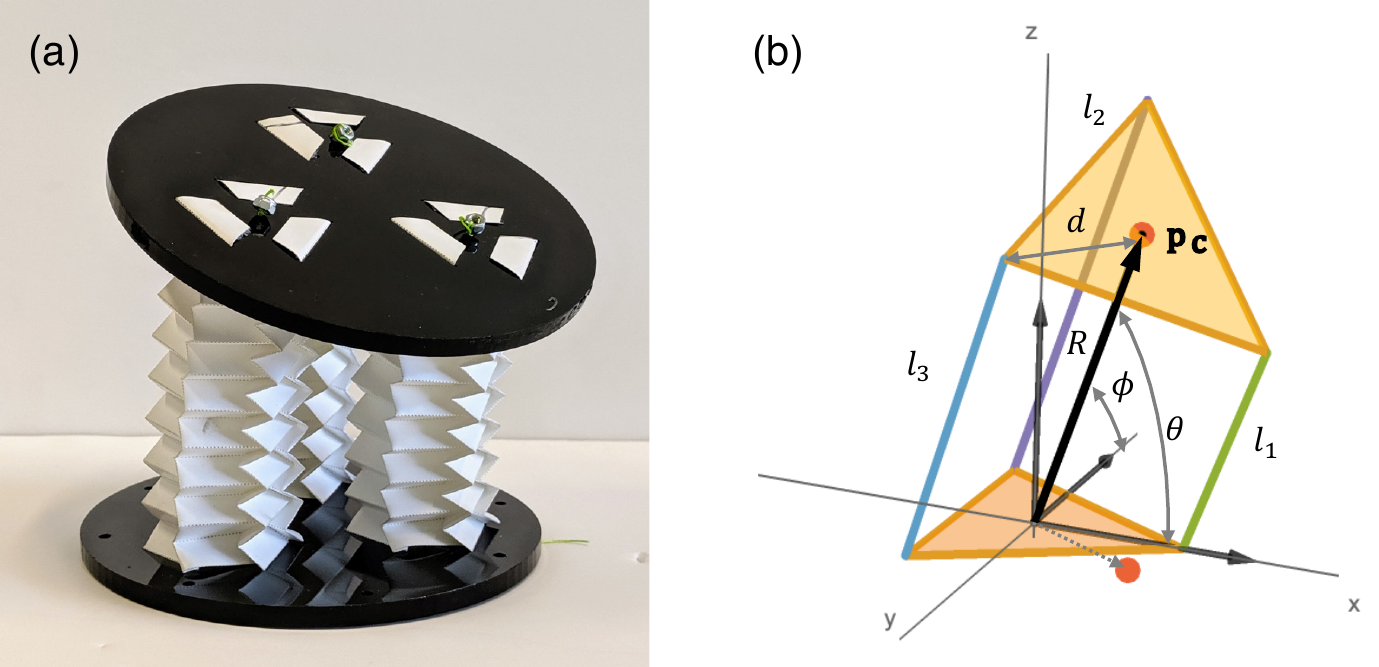}
      \caption{(a) The compliant body composed of REBO. (b) Kinematic model of the REBO Juggler's tendon driven top plate under compressive load}
      \label{fig4}
    \end{figure}

%This work proposed a linear actuator composed of a REBO structure as a linear spring, with an inner tendon to be its antagonistic pair. The compliant body is composed of three parallel linear actuator arranged into a truncated triangular prism structure, and could be modeled as Fig. \ref{fig4}(b). The top and the bottom rigid plates are congruent regular triangles, with the linear actuators mounted on the vertices of the triangles. Controlling the length of the tendon results in changes of the state of the top plate. Due to geometric symmetry, each joining face of the truncated prism is an isosceles trapezoids.

The robot platform can be modeled as two equilateral triangles connected at the corners by three tendons (Fig.~\ref{fig4}(b)), where the change of tendon length changes the position and orientation of the top triangle.
Let the origin of the model be at the center of the bottom triangle. The top and the bottom triangles have circumcircles  of radius $d$. The linear actuator state is defined as $q_{la}=(l_1, l_2, l_3)\in [l_{min}, l_{max}]^3=Q_{la}$, where $l_1$, $l_2$, and $l_3$ are the length of the three parallel linear actuators, and $l_{min}$, $l_{max}$ are the length constraints of the actuators. The tendon is attached to a pulley mounted on the motor, where the motor state is defined as $q_{m}=(\theta_{m,1}, \theta_{m,2}, \theta_{m,3})\in \mathbb{T}^3=Q_{m}$, and the mapping from the motor space to the linear actuator space $q_{la}=f_1(q_m)$ is: 
    \begin{equation}
        l_i = l_0+r_p \theta_{m,i}
    \end{equation}
where the index $i$ indicates different actuator pairs, $l_0=l_{max}$ is the rest length of REBO, $r_p$ is the radius of the pulley and $\theta_{m,i}$ is the angle of rotation of the corresponding motor.

The position vector of the center of the top triangle is $\mathbf{p_c}$. For this three DOF system, the orientation of the top triangle is coupled with its position, which can be fully described with $q_{tt}=(r, \theta, \phi)\in[l_{min},l_{max}]\times\mathbb{S}^1\times\mathbb{S}^1=Q_{tt}$, where $r$ is the length of $\mathbf{p_c}$, $\theta$ is the angle between $\mathbf{p_c}$ and the $x$ axis, and $\phi$ is the angle between $\mathbf{p_c}$ and the $y$ axis. The kinematics $q_{tt}=f_2(q_{la})$ can found by observing the geometry of the model to be:
    \begin{align}
    r &= \tfrac{1}{3}(l_1+l_2+l_3)
    \label{eq:fk}\\
    \theta &= \cos^{-1}\left(\tfrac{1}{6 d}(-2l_1+l_2+l_3)\right)\\
    \phi &= \cos^{-1}\left(\tfrac{1}{2\sqrt{3} d}(-l_2+l_3)\right).
    \end{align}
The position vector $\mathbf{p_c}=\begin{bmatrix}x_c&y_c&z_c\end{bmatrix}^T\in\mathbb{R}^3$ can be described in Cartesian coordinates as
    \begin{equation}
    \begin{split}
    \mathbf{p_c}&= f_{cc}(q_{tt})\\
    :&= \begin{bmatrix}r \cos{\theta}&r \cos{\phi}&r\sqrt{1-\cos^2{\theta}-\cos^2{\phi}}\end{bmatrix}^T .
    \end{split}
    \end{equation}
When controlling the robot, the input command to the motor can be found by the inverse map of the kinematics as
    \begin{equation}
        q_m=f_1^{-1}\circ f_2^{-1}\circ f_{cc}^{-1}(\mathbf{p_c}).
    \label{eqINV}
    \end{equation}
    
% \subsection{Tendon driven control for different stages}

\section{Kinematic Task: Pointing}
    \begin{figure}[t]
      \centering
      \includegraphics[scale=0.58]{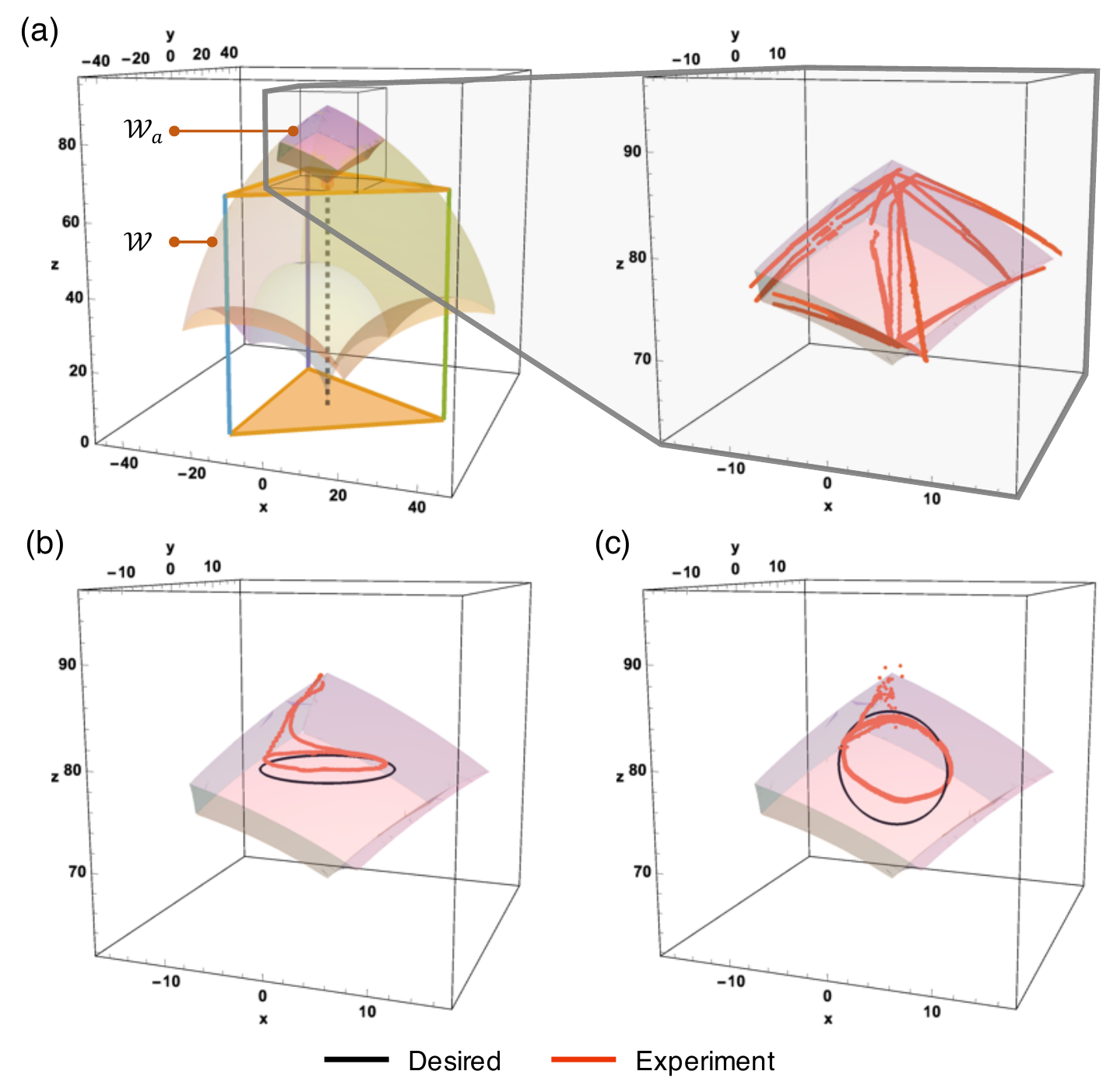}
      \caption{(a) The kinematically achievable workspace $\mathcal{W}$ and the actuator achievable workspace $\mathcal{W}_a$ of the REBO Juggler's Top plate. The colored bent cuboid is the simulation where the red line is the experimental trajectories of the envelope of the workspace. (b) horizontal circle test (c) vertical circle test.}
      \label{fig5}
    \end{figure}
The trajectory of the REBO Juggler's top plate can be planned within its constrained workspace, and the controlling command of the motor can be found through the inverse map as shown in Eq.~\eqref{eqINV}. The kinematically achievable workspace of the top plate is the image through the kinematic model of the extreme of the length of the tendon $l_{max}$ and $l_{min}$ and can be described as $\mathcal{W}=\{\mathbf{p_c}|\mathbf{p_c}=f_{cc}\circ f_2(q_{la}), q_{la}\in Q_{la}\}$. The actuator achievable workspace $\mathcal{W}_a$ is a subset of the full workspace $\mathcal{W}$, and is limited by the continuous torque $\tau_c$ of the motor, where its minimum value can be found as $l_{min}=l_{max}-(\tau_c/r_p )/K_s$, where $K_s$ is the effective stiffness of one REBO. Since this research explores the structure with high stiffness, the achieved compression $=l_{max}-l_{min}$ is small. Fig. \ref{fig5} shows the image through the kinematic model $\mathcal{W}$, with $l_{max}=\SI{88}{\milli\metre}$ and $l_{min}=\SI{66}{\milli\metre}$. The constrained workspace $\mathcal{W}_a$ has a volume of $\SI{4980.95}{\milli\metre^3}$. 

To demonstrate the mobility of the top plate, we tracked the position of the top plate under varying control inputs using an OptiTrack motion capture system. The first experiment was a open-loop workspace boundary test. It can be seen from Eq.~\eqref{eq:fk} that the boundary of the workspace occurs where two of the linear actuators are at their maximum (or minimum) length and the third changes length. We therefore measured the trajectory of the top plate when each linear actuator was moved between $l_{min}$ and $l_{max}$ individually while holding the other constant. The experimental results are shown as the red trajectories in Fig. \ref{fig5}(a), which capture the structure of the simulated workspace, and has a $14.4\%$ rms error from the predicted boundary. The volume of the convex hull of these trajectories is $\SI{5426.26}{\milli\metre^3}$, which has a $8.94\%$ error from the predicted volume.

To check the accuracy of the kinematic model, we commanded the top plate of the robot to follow  circular trajectories that were generated to lie within the workspace. The top plate was set to follow first a horizontal circle with radius $\SI{7}{\milli\metre}$, then a vertical circle with radius $\SI{6}{\milli\metre}$, both centered at $\begin{bmatrix}0&0&\tfrac{1}{2}(l_{min}+l_{max})\end{bmatrix}^T$. Fig. \ref{fig5}(b) and \ref{fig5}(c) show the results of the two experiments. The root mean square errors for the two tracking tests were $11.32\%$ and $14.66\%$, respectively. Fig. \ref{fig5}(c) shows that the real and desired trajectories of the vertical circle deviate closer to the bottom. This is because  the actuator needs more torque for greater compression, yet is limited by the motors' maximum continuous torque. This limitation is expected to be eliminated when we change to a motor with a greater torque or have a higher bandwidth controller.
% \WC{Cynthia you are right, it should have been enough, however, enough continuous torque on the hardware side doesn't mean we have the capability to do so, (1) the PD gain we choose are not optimal, and there are no i gain, thus there is a high chance of offset and error. (2) the position control in Ghost SDK is a black box, i.e., there's another control loop that has a PID gain that I don't know. So I think this is could be an control issue that i did not try to improve.}\CS{Alright, I guess this is one of those things we keep, and then think about fixing for the camera ready...}

\section{Dynamical Task: Vertical Juggling}
% \subsection{Vertical Juggler}
    % \begin{algorithm}[t]
    % %  \SetInd{0.5em}{0.5em}
    %  \KwIn{$\chi_{pp}$}
    %  \KwOut{$h_{ball}$}
    %  \Begin{
    %  $mode \leftarrow Flight$\;
    %     \Switch{mode}{
    %       \Case{flight}{
    %         $\chi \leftarrow \chi_{pp}$\;
    %         $mode \leftarrow (force = 1 ? Hit: Flight)$\;
    %       }
    %       \Case{Hit}{
    %         $\chi \leftarrow \chi_0$\;
    %         $mode \leftarrow (force = 0 ? Flight: Hit)$\;
    %       }
    %     }
    %   }
    %   \caption{Continuous ball juggling}
    % \end{algorithm}
    \begin{figure}[t]
      \centering
      \includegraphics[scale=0.58]{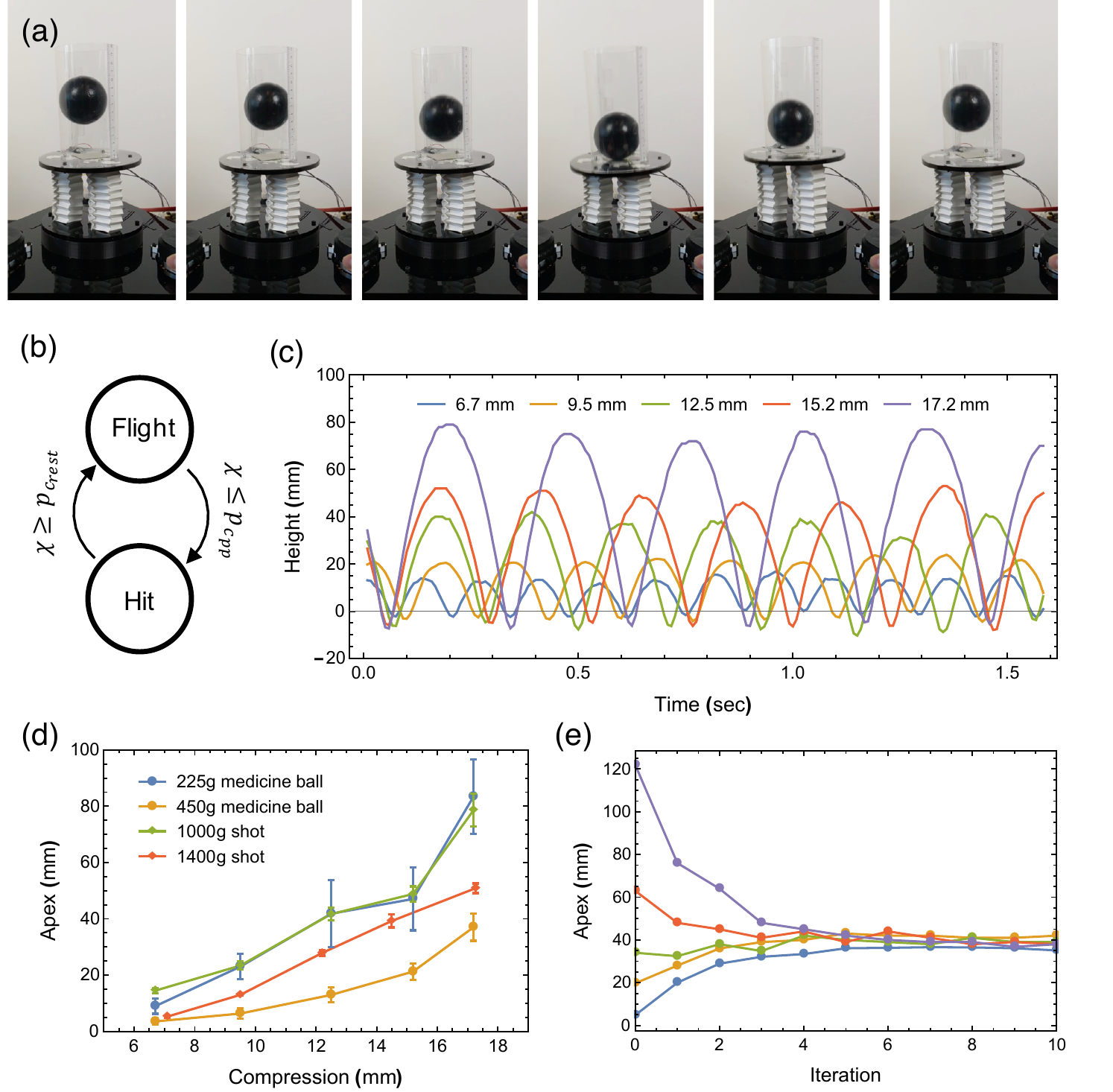}
      \caption{Vertical juggling (a) Snapshot of the experiment with an interval of 1/30 sec. (b) State machine of the juggling task. (c) Trajectory of the 1 kg shot under different pre-pressed REBO conditions. (d) Pre-pressed length of REBO v.s. juggling height of several different balls with STDs. (e) Transient responses (apex vs. iteration) of the 1 kg shot from a variety of initial heights, exhibiting the asymptotically stable fixed point at 4 cm associated with the 12.5 mm pre-compression setting plotted in (d). }
      \label{fig6}
    \end{figure}

% To demonstrate that REBO-mediated actuation is able to perform dynamical tasks, a juggling task is assigned to the REBO Juggler, where it is required to repeatedly hit a weighted ball into the air with its top plate as the paddle. Stable juggling is a quintessentially dynamical task in the sense of requiring carefully controlled exchanges of potential and kinetic energy between a robot and its environment \cite{buhler1990family}. The task is an effectively simple but appropriately challenging precursor to more general dynamically dexterous capabilities because the duration of travel during paddle-ball contact is small while the force required to change the direction of the motion of the ball is substantial.
Reflecting the past traditions of the field of dynamical locomotion, the core problem of stable running \cite{raibert1986legged} can be reduced to the problem of stable vertical hopping \cite{raibert1983dynamic}, for which a purely vertical juggle is a representative surrogate \cite{buehler1990simple,koditschek1991analysis}. Thus, as a proof-of-concept task, the REBO juggler's ball, its dynamical 
"environment," is restricted to purely vertical motion by confinement within a tube mounted on the paddle, while the  parallel three REBO elements can be viewed as an actively loaded single virtual vertical spring. Since the REBO exhibits elastic hysteresis, the total energy of the ball would be substantially diminished by each encounter with a relaxed paddle,  hence additional energy must be pre-loaded into the waiting spring so as to impart at each hit the work needed to keep the ball bouncing. Specifically, to pump energy into the coupled robot-environment system, the motors work on the REBO by pre-compressing it to a fixed position $\mathbf{p_{c_{p}}}$, before the ball lands on the paddle. When the sensing pad detects contact, the REBO then releases the energy into the ball by resetting the set point of the tendon to its rest length $\mathbf{p_{c_{r}}}$. Fig. \ref{fig6}(a) shows a successful juggling period of one of the open-loop experiments presented  in this section.

Juggling arises from a hybrid dynamical system comprising two modes: ``Flight" and ``Hit", as shown in Fig. \ref{fig6}(b). In the ``Flight" mode, the launched ball  exhibits a ballistic trajectory governed by the lossless constant gravity system ${m_b\ddot{\chi}= -m_b g}$, where $m_b$ and $\chi$ are the mass and the position of the ball, respectively. After the ball's launch, the juggler quickly resets back to its pre-compressed position. Because the compliant REBO structure has a mass that is negligible (less by at least an order of magnitude) relative to that of the frame or ball, we ignore any REBO dynamics and treat the reset as instantaneous. The ``Flight" mode stops when the force sensor on the paddle is triggered by the ball's contact and the system enters the ``Hit" mode. Now, the ball rides down the paddle of the compressing REBO structure, and the whole system can be considered as a mass on a spring. The REBO elastic energy is imparted to the ball's mass as governed by the dynamics $m_b\ddot{\chi}= -K_{es}(\chi-r_{rest})-B_s\dot{\chi}-m_b g$, where $K_{es}=3K_s$ is the juggler's effective stiffness  and $B_s$ its damping coefficient. The ``Hit" mode ends when the REBO reaches its rest length, whereupon the ball lifts off as reported by the force sensor, the motor re-engages the tendon, and the system re-enters the ``Flight" mode.

REBO's Hookean force-extension curve (Fig. \ref{fig2}(c)) implies that the more it is compressed, the more energy it stores; hence, because it can sustain high forces under load, the juggler injects more energy into the ball in ``Hit" mode with greater pre-compression, resulting in higher apex positions. Fig. \ref{fig6}(c) documents this increase in vertical oscillation amplitude for a \SI{1}{kg} shot under increasing commanded pre-compression lengths $\mathbf{p_{com}}=\mathbf{p_{c_{r}}}-\mathbf{p_{c_{p}}}$. A slow motion video of 120 fps has been filmed for every trial and the trajectory of the ball was found using ``Tracker" (https://physlets.org/tracker/). The result confirms that the more the REBO is pre-compressed, the higher the ball can be juggled. 

Fig. \ref{fig6}(d) summarizes the results of  repeated (100 juggling cycles each)  experiments with several different balls being juggled under different pre-compressed lengths by plotting the mean and standard deviation of measured apex heights. Two shots of mass \SI{1}{kg} and \SI{1.4}{kg} were selected because their resilient rigid metal composition yields an approximately elastic collision,  presenting  a ``lossless environment" to the juggler. In contrast, two sand-loaded medicine balls of mass \SI{225}{g} and \SI{450}{g} yield highly inelastic collisions chosen to confront the juggler with a ``highly dissipative environment." It is clear that the average apex height is monotonically increasing with respect to the pre-compression for all the balls. For the same compressive pre-load condition, the heavier balls have a lower apex than the lighter ones as expected if each pair is restored to the same steady energy state. Here, since the medicine balls dissipate more stored energy than the shots, making them harder to juggle, the juggler lofts the \SI{225}{g} ball to roughly the same steady state apex as the \SI{1000}{g} shot. We can summarize  the energetic properties of the REBO structure with respect to its work on the balls as follows. The energy loaded  into the REBO structure by the DC servos' pre-compression work is  $E=\tfrac{1}{2} K_{es} |\mathbf{p_{com}}|^2 =\tfrac{1}{2}(3105)*(0.0172)^2\approx \SI{0.5}{J}$ for the \SI{1}{kg} shot bounced at a height of \SI{8}{cm}. Since the ``Hit" mode has a typical duration of \SI{0.02}{s} the REBO delivers a mechanical power output of  \SI{25}{W}.

%\WC{Dan, I need your help writing this paragraph about the convergence map}. How stable is the REBO juggler? 
Fig. \ref{fig6}(e) plots the trajectories over the course of the first ten successive collisions (out of hundreds recorded) with the juggler's paddle of the apex heights of the \SI{1}{kg} shot starting from five different initial conditions, all subject to the same pre-compressed REBO length of \SI{12.5}{\milli\metre}. Treating the apex height as the coordinate chart for the Poincar\'{e} section of this hybrid dynamical system, the plot demonstrates the asymptotic stability of the period one hybrid limit cycle by displaying convergence to the \SI{4}{cm} fixed point of the associated Poincar\'{e} (or ``return") map \cite{buehler1990simple}.   The  results suggest the relatively large basin of attraction (set of initial heights that are successfully juggled up or down to the desired \SI{4}{cm} steady state apex height)  achieved by the juggler consistent with a high power actuator along the lines discussed in \cite{koditschek1991analysis}.

\section{Discussion and Future Work}

The resilient, tunable stiffness of the lightweight, deformable REBO structure allows us to transfer energy through  the 1kg shot at roughly \SI{25}{W},  repeatedly over the course of thousands of hits with very little fatigue, as attested by the highly repeatable asymptotically stable steady state juggling cycles, lofting 1 kg loads to nearly 10 cm heights. 
%the desired force we need with given deformation, and with small deformation, it can generate the force to push the shot up very fast, thus providing high power exchange. The juggling experiment shows that the resilience of REBO is strong, and the stored elastic energy can be turned into kinetic energy repeatedly and steadily. The nearly 3000 total juggling cycles throughout the experiments show no fatigue, highlighting the resilience of REBO. 
Thus, the REBO breaks new ground in the soft robotics literature by transducing energy, $25/0.25 \approx$ \SI{100}{(\watt\per\kilogram)} which, when distributed across the repeating origami structure, is sufficient to power the task of vertical juggling ---  an  established route to dynamical dexterity in conventional robotics \cite{buehler1990simple}.  Indeed, we have begun work to turn the REBO juggler ``upside down," aiming for a power autonomous ``soft" hopper capable of lifting its batteries and actuators to comparable apex states.     
The experiments presented here focus on merely modifying the cone angle $\beta$ for REBOs made of the same materials with the same thicknesses and having the same number of sides and side lengths. Future work yielding a more formal understanding of the REBO's properties will afford  scaling laws that achieve a generalizable tunably compliant design methodology for a broad range of robotics applications.

% \CS{Can you add one sentence here on these challenges (i.e., what is the work required relative to the results you have in this paper), then one sentence connecting the potential impact of the resulting platform back to the vision in the first paragraph?}
%\WC{Carrying its own weight and trying to balance itself while hopping has been a major challenge in rigid robots. A stable hopping \cite{koditschek1991analysis} would show that the new origami based meta-material concept presented here could offer a powerful route to create stable running \cite{raibert1983dynamic} robots. }

%\WC{Dan, the weights is in section III. A. each REBO is 16g, with the acrylic is 220g}
% \Kod{Wei-Hsi, it would be nice to express these absolute heights in terms of specific agility numbers \cite{Duperret_Kenneally_Pusey_Koditschek_2016} for a hypothetical 1 kg hopping robot.}\WC{Sure, I'm on it}

\addtolength{\textheight}{-3cm}   % This command serves to balance the column lengths

%%%%%%%%%%%%%%%%%%%%%%%%%%%%%%%%%%%%%%%%%%%%%%%%%%%%%%%%%%%%%%%%%%%%%%%%%%%%%%%%

%%%%%%%%%%%%%%%%%%%%%%%%%%%%%%%%%%%%%%%%%%%%%%%%%%%%%%%%%%%%%%%%%%%%%%%%%%%%%%%%

%%%%%%%%%%%%%%%%%%%%%%%%%%%%%%%%%%%%%%%%%%%%%%%%%%%%%%%%%%%%%%%%%%%%%%%%%%%%%%%%
%\section*{APPENDIX}

%Perhaps the derivation of finding $\alpha$ given $\beta$

\section*{Acknowledgment}

 This work is supported in part by the Army Research Office (ARO) under the SLICE Multidisciplinary University Research Initiatives (MURI) Program, award \# W911NF1810327.  We thank Diego Caporale for technical consultant and Diedra Krieger for administrative support.

%%%%%%%%%%%%%%%%%%%%%%%%%%%%%%%%%%%%%%%%%%%%%%%%%%%%%%%%%%%%%%%%%%%%%%%%%%%%%%%%

\bibliographystyle{IEEEtran}
\bibliography{refs}

\end{document}